\documentclass[letterpaper, 10 pt, conference]{ieeeconf}  
\let\labelindent\relax
\usepackage{enumitem}
\usepackage[utf8]{inputenc}

\IEEEoverridecommandlockouts                              

\usepackage{cite}
\usepackage{amsmath,amssymb,amsfonts}
\usepackage{dsfont}
\usepackage{bbm}

\usepackage{graphicx}
\usepackage{textcomp}
\usepackage{cuted}
\usepackage{capt-of}
\usepackage{bibunits}
\usepackage{float}
\usepackage{booktabs}
\usepackage{threeparttable}
\usepackage{mathtools}
\usepackage[mathscr]{eucal}
\usepackage[table,xcdraw]{xcolor}
\usepackage{colortbl}
\usepackage[utf8]{inputenc}
\usepackage{colortbl}
\usepackage{balance}
\usepackage{color}
\usepackage{multirow}
\usepackage{caption}
\usepackage{marvosym}
\usepackage{indentfirst}
\usepackage{makecell}
\usepackage{gensymb}
\usepackage{nicematrix}
\usepackage[table]{xcolor}
\definecolor{car}{rgb}{0.39215686, 0.58823529, 0.96078431}
\definecolor{bicycle}{rgb}{0.39215686, 0.90196078, 0.96078431}
\definecolor{motorcycle}{rgb}{0.11764706, 0.23529412, 0.58823529}
\definecolor{truck}{rgb}{0.31372549, 0.11764706, 0.70588235}
\definecolor{other-vehicle}{rgb}{0.39215686, 0.31372549, 0.98039216}
\definecolor{person}{rgb}{1.        , 0.11764706, 0.11764706}
\definecolor{bicyclist}{rgb}{1.        , 0.15686275, 0.78431373}
\definecolor{motorcyclist}{rgb}{0.58823529, 0.11764706, 0.35294118}
\definecolor{road}{rgb}{1.        , 0.        , 1.        }
\definecolor{parking}{rgb}{1.        , 0.58823529, 1.        }
\definecolor{sidewalk}{rgb}{0.29411765, 0.        , 0.29411765}
\definecolor{other-ground}{rgb}{0.68627451, 0.        , 0.29411765}
\definecolor{building}{rgb}{1.        , 0.78431373, 0.        }
\definecolor{fence}{rgb}{1.        , 0.47058824, 0.19607843}
\definecolor{vegetation}{rgb}{0.        , 0.68627451, 0.        }
\definecolor{trunk}{rgb}{0.52941176, 0.23529412, 0.        }
\definecolor{terrain}{rgb}{0.58823529, 0.94117647, 0.31372549}
\definecolor{pole}{rgb}{1.        , 0.94117647, 0.58823529}
\definecolor{traffic-sign}{rgb}{1.        , 0.        , 0.    }

\makeatletter
\usepackage{tikz}
\usepackage{lipsum}
\usepackage{caption}
\usepackage{hyperref}
\newcommand*\circled[1]{\tikz[baseline=(char.base)]{
            \node[shape=circle,draw,inner sep=0.5pt] (char) {#1};}}
\let\NAT@parse\undefined
\makeatother

\usepackage{enumitem}

\title{\LARGE \bf
Collaborative Perceiver: Elevating Vision-based 3D Object Detection \\ via Local Density-Aware Spatial Occupancy
}
\author{Jicheng Yuan$^{1}$, Manh Nguyen Duc$^{1}$, Qian Liu$^{1,2}$, Manfred Hauswirth$^{1,2}$ and Danh Le Phuoc$^{1,2}$
\thanks{$^{1}$The authors are with the Open Distributed Systems (ODS) Group at the Technische Universität Berlin and $^{2}$ Fraunhofer FOKUS (Berlin, Germany). \{firstname.lastname@tu-berlin.de\}}%
}

\begin{document}

\maketitle
\thispagestyle{empty}
\pagestyle{empty}

\begin{abstract}
Vision-based bird's-eye-view (BEV) 3D object detection has advanced significantly in autonomous driving by offering cost-effectiveness and rich contextual information. However, existing methods often construct BEV representations by collapsing extracted object features, neglecting intrinsic environmental contexts, such as roads and pavements. This hinders detectors from comprehensively perceiving the characteristics of the physical world. To alleviate this, we introduce a multi-task learning framework, Collaborative Perceiver (CoP), that leverages spatial occupancy as auxiliary information to mine consistent structural and conceptual similarities shared between 3D object detection and occupancy prediction tasks, bridging gaps in spatial representations and feature refinement. To this end, we first propose a pipeline to generate dense occupancy ground truths incorporating local density information (LDO) for reconstructing detailed environmental information. Next, we employ a voxel-height-guided sampling (VHS) strategy to distill fine-grained local features according to distinct object properties. Furthermore, we develop a global-local collaborative feature fusion (CFF) module that seamlessly integrates complementary knowledge between both tasks, thus composing more robust BEV representations. Extensive experiments on the nuScenes benchmark demonstrate that CoP outperforms existing vision-based frameworks, achieving 49.5\% mAP and 59.2\% NDS on the test set. Code is available at this \href{https://github.com/jichengyuan/Collaborative-Perceiver}{link}.

\end{abstract}

\section{INTRODUCTION}
3D object detection (OD)~\cite{feng2023aedet,mao20233d} has been a vital component in autonomous driving (AD) and mobile robots. Compared to LiDAR, cameras can provide richer contextual information regarding object semantics and offer the advantage of cost-effectiveness. One practical paradigm utilizes multi-view images as input to identify and localize objects of interest~\cite{huang2021bevdet,li2023bevdepth,huang2022bevdet4d,li2022bevformer,liu2023petrv2}. For instance,  BEVDet~\cite{huang2021bevdet} transforms 2D multi-view image features into 3D space using the Lift-Splat-Shoot (LSS) based view transformation~\cite{philion2020lift}. Then, these transformed spatial features are collapsed into BEV grids along the height dimension, upon which detection decoders are applied. Building on BEVDet, many studies~\cite{li2023bevdepth,huang2022bevdet4d,li2022bevformer} further extended it with depth supervision~\cite{li2023bevdepth}, temporal fusion~\cite{huang2022bevdet4d}, and spatial-temporal fusion~\cite{li2022bevformer}, leading to significant improvements in 3D perception.
\begin{figure}[t]
\begin{center}
\includegraphics[width=0.9\linewidth]{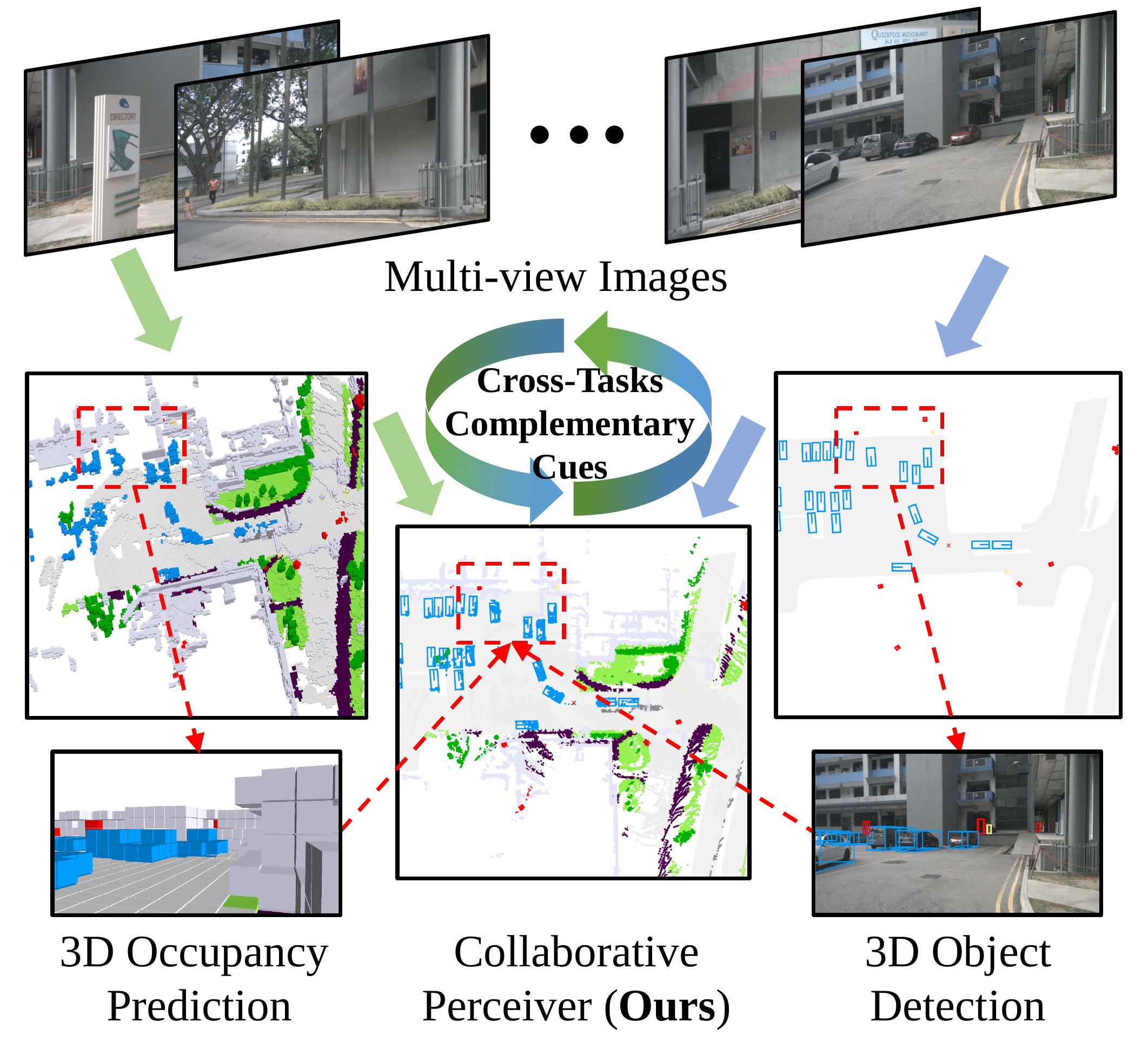}
\end{center}
\setlength{\abovecaptionskip}{1pt}
\caption{
Overview of the proposed \textbf{Co}llaborative \textbf{P}erceiver (CoP), which leverages complementary and consistent knowledge from 3D occupancy and 3D object detection to achieve comprehensive scene understanding.
}
\vspace{-7mm}
\label{fig:fig1}
\end{figure}

Despite the advancements of the aforementioned methods~\cite{huang2021bevdet,li2022bevformer,li2023bevdepth,huang2022bevdet4d}, there are not without challenges: \circled{C1} Existing 3D detectors~\cite{mao20233d} struggle to identify objects with unique or irregular geometries, making it difficult to ensure the safety of AD under complex traffic scenarios~\cite{hauer2019did}. In contrast, 3D occupancy (Occ.) describes the physical world by assigning an occupied probability to each voxel in the 3D space, delivering a geometrically consistent and precise description of dynamic scenes. However, many existing
methods~\cite{wei2023surroundocc,yu2023flashocc} inherently assume a homogeneous point density during voxelization, potentially overlooking
voxel-level point density variations that are essential for representing fine-grained object properties. \circled{C2} Traditional BEV-based approaches~\cite{huang2021bevdet,huang2022bevdet4d,li2023bevdepth,li2022bevformer} collapse multi-view features along the height dimension, constructing flattened BEV features. However, identifying objects with distinct spatial structures in BEV space is challenging if the dimensionality compression step cannot adaptively preserve their spatial properties~\cite{liu2023bevfusion}, e.g., a traffic cone at a low height versus a truck at a higher one, as demonstrated in Figure~\ref{fig:fig1}. \circled{C3} Relying solely on task-specific local or global information hampers the construction of a universal environmental representation, potentially resulting in inconsistent target recognition and suboptimal performance~\cite{keetha2023anyloc}. Thus, it is crucial to efficiently extract complementary and consistent knowledge from both structured and unstructured elements of the scene, achieving an effective and comprehensive perception of dynamic environments.

In this work, we propose \textbf{Co}llaborative \textbf{P}erceivers (\textbf{CoP}), a multi-task learning framework that seamlessly integrates complementary global and local cues shared between 3D object detection and occupancy prediction to build robust BEV representations, thereby boosting 3D detection performance. To tackle challenge~\circled{C1}, \textbf{CoP} employs 3D occupancy prediction as an auxiliary task to capture informative spatial local features and achieve fine-grained geometric consistency. Considering the significant effort often required to obtain dense occupancy supervision and the non-uniform distribution of point clouds during voxelization, we propose a pipeline that automatically generates local density-aware dense occupancy (LDO) ground truths from offline multi-frame LiDAR data, enriching environmental detail. To alleviate challenge~\circled{C2}, \textbf{CoP} employs a voxel-height-guided sampling (VHS) strategy to extract height-aware local features, aiming to iteratively refine learned latent spatial information across distinct height ranges. Building upon this, we further develop a collaborative feature fusion (CFF) module that leverages the synergistic cues from global and local knowledge throughout the learning process, thereby addressing challenge~\circled {C3}. By establishing consistent knowledge shared between tasks, CoP aims to understand environmental correspondences from joint perspectives and enhance perception of structural integrity in the physical world. \\ In summary, our contributions are as follows:
\begin{itemize}[leftmargin=*, itemsep=0pt] 
\item {We introduce spatial occupancy states as auxiliary information that, together with LDO ground truth, reconstruct fine-grained spatial structures and enrich environmental detail, thereby boosting 3D detection.}
\item {We propose a VHS module that leverages vertical positioning information as a prior to capture fine-grained spatial features across diverse object properties.}
\item {Building upon VHS, our CFF module further mines structural correspondences, facilitating feature extraction, enhancement, and interaction from combined global and local perspectives, yielding robust BEV representations.} 
\item {Extensive experiments on nuScenes benchmark~\cite{caesar2020nuscenes} demonstrate that CoP achieves competitive performance against vision-based frameworks, notably with $+$2.1\% mAP and $+$1.1\% NDS improvements, consistently enhancing 3D object detection through an effective learning strategy.}
\end{itemize}

\section{Related Work}
\begin{figure*}[ht]
  \centering
  \includegraphics[width=0.9\textwidth, 
                ]{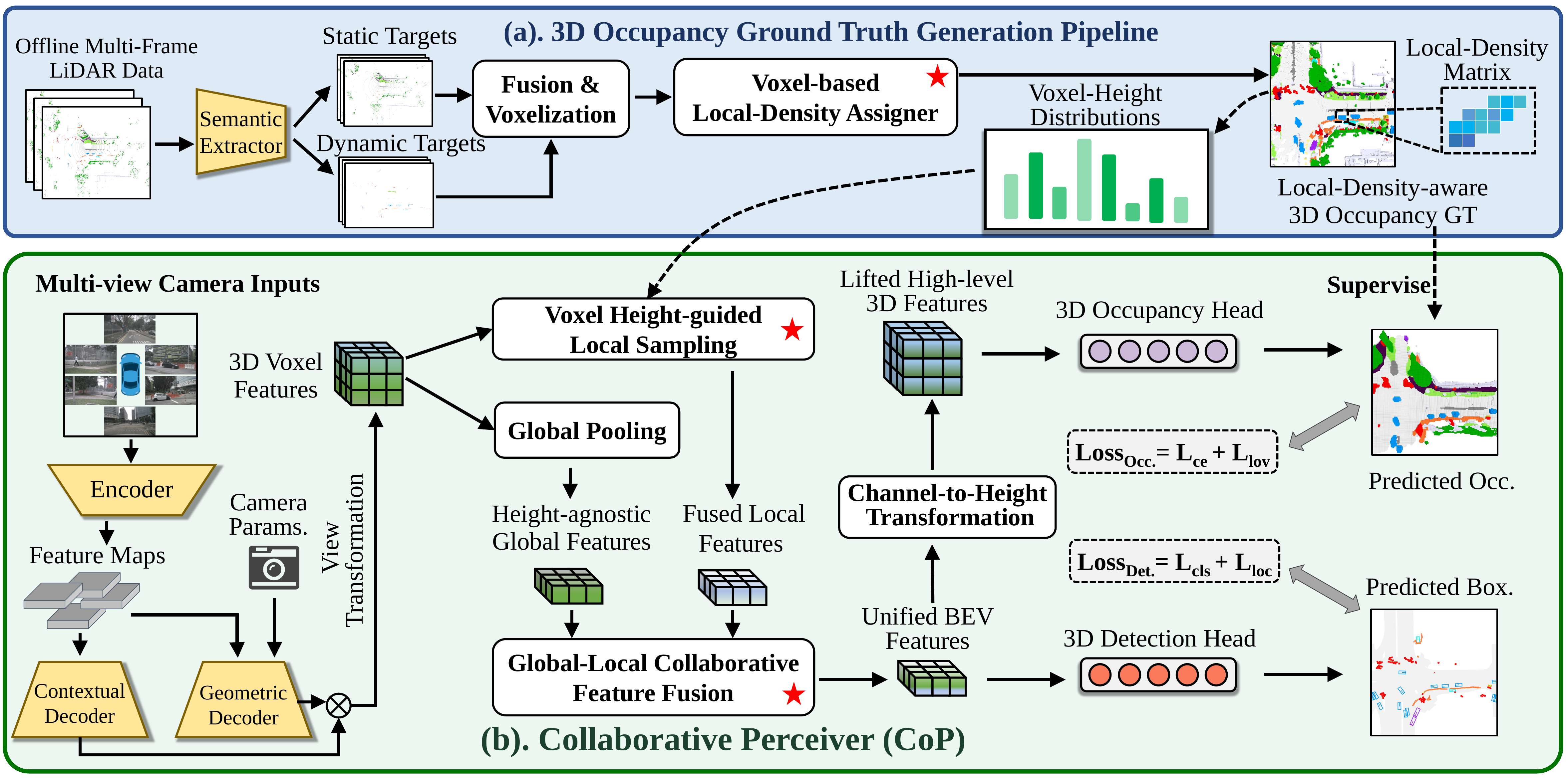} 
  \caption{Overview of the proposed \textbf{Co}nsistent \textbf{P}erceiver (CoP) and 3D occupancy ground truth (GT) generation pipeline. First, multi-view image features are extracted and transformed into the 3D space following the LSS~\cite{philion2020lift}. Then, global pooling and local sampling are employed to extract coarse global and fine-grained local features guided by the voxel-height distributions. Next, a global-local collaborative feature fusion block with a C2H module enables adaptive feature extraction, enhancement, and interaction between these features. In addition, the offline-generated local-density-aware 3D occupancy GT with informative and detailed scene semantics works as auxiliary supervision.}
  \label{fig:overview}
\end{figure*}
\noindent \textbf{Vision-centric 3D Object Detection}  aims to identify 3D objects from images captured by single or multiple cameras\cite{wang2022performance}. Prior methods~\cite{zhang2021objects, park2021pseudo,brazil2019m3d,ma2021delving,chen2016monocular} extend advanced 2D object detection methods~\cite{tian2020fcos,duan2019centernet} to the 3D domain by inferring 3D attributes from 2D counterparts, e.g., FCOS3D~\cite{wang2021fcos3d} capitalizes on the strong spatial correlation between object's characteristics and their visual representations to discern 3D objects. Subsequently, BEV-based methods~\cite{chi2023bev,xie2204m2bev, liu2023bevfusion} have emerged as the mainstream approach. For instance, BEVDet~\cite{huang2021bevdet} leverages the Lift-Splat-Shoot (LSS)~\cite{philion2020lift} view transformation to convert images from surround-view cameras into unified BEV features for full-range detection. Then, BEVDepth~\cite{li2023bevdepth} augments BEVDet~\cite{huang2021bevdet} with a camera-aware depth network for more accurate depth estimation, while BEVDet4D~\cite{huang2022bevdet4d} expands it into the spatiotemporal domain. Meanwhile,  a range of query-based methods~\cite{wang2022detr3d,liu2022petr,liu2023sparsebev} strive to integrate DETR~\cite{carion2020end} into the 3D domain. Recent advancements have further enhanced 3D detection by leveraging long-term temporal fusion~\cite{liu2022petr} and sparse representations~\cite{liu2023sparsebev}. However, identifying objects exhibiting irregularly spatial structures in flattened BEV space without detailed object properties, especially in complex traffic environments~\cite{hauer2019did}, remains challenging for 3D detectors.
\\
\textbf{Vision-centric 3D Occupancy Prediction} seeks to estimate detailed environmental information from images, such as semantic occupancy (Occ.) states at positions within 3D scenes, thereby facilitating downstream planning~\cite{ho2018virtual,hrabar20083d} and navigation tasks~\cite{elfes1989using,ramakrishnan2020occupancy}. Its earliest origins can be traced back to Occupancy Grid Maps (OGM)\cite{wang2022probabilistic,moravec1985high} navigating mobile robots within static scenes. One step forward, TPVFormer~\cite{huang2023tri} extends BEV to tri-perspectives in forecasting 3D occupancy. However, its output is sparse due to limited LiDAR data. Subsequently, SurroundOcc~\cite{wei2023surroundocc} and Occ3D~\cite{tian2024occ3d} proposed to convert sparse LiDAR sequences into dense semantic occupancy, yielding informative representations. OccNeRF~\cite{zhang2023occnerf} further introduces an occupancy prediction network without 3D supervision, advancing 3D occupancy prediction in LiDAR-free environments. Concurrently, SparseOcc~\cite{tang2024sparseocc} rethinks sparse latent representation for vision-based semantic occupancy prediction, inspired by efficient occupancy networks and sparse point cloud processing~\cite{yang20203dssd, yang2019std}. 
However, these solutions assume a homogeneous point density distribution during voxelization, overlooking the local density as a critical factor in representing informative object properties. Given the inherent limitations in describing objects with irregular spatial structures using bounding boxes and the advanced semantic scene representation capacity of 3D occupancy, we argue that one unified framework combining shared cues and complementary knowledge from both tasks can be a promising solution for achieving a more comprehensive understanding of dynamic scenes. 
\section{Collaborative Perceiver}
\subsection{Overview and Problem Formulation} \label{subsec:overview}
Despite significant strides in multi-camera-based 3D detection~\cite{wang2022performance} achieved
by previous methods~\cite{huang2021bevdet, li2022bevformer, li2023bevdepth,liu2023bevfusion,huang2022bevdet4d}, solely relying on box-level descriptions cannot provide a holistic perception of 3D scenes, which is crucial for ensuring the safety of autonomous vehicles~\cite{arnold2019survey}. To this end, different from prior approaches~\cite{huang2021bevdet, li2022bevformer, tian2024occ3d, zhang2023occformer}, we propose a unified collaborative learning framework, CoP, to capture complementary cues and structural similarities from a joint perspective of object detection and occupancy prediction. 

Initially, defining multi-camera images $I\in\mathbb{R}^{N \times W_I \times H_I \times 3}$ as inputs, where $N$ is the number of surrounding cameras, we concatenate ResNet~\cite{he2016deep} and FPN~\cite{lin2017feature} as our image encoder, to extract multi-view image features $f_{i} \in \mathbb{R}^{N \times C_i \times H_i \times W_i}$, where $H_i$, $W_i$ are downscaled to $1/16$ of the original $H_I$, $W_I$ dimensions, as depicted in Figure~\ref{fig:overview}(b). Then, we employ Lift-Spalt-Shoot (LSS)~\cite{philion2020lift} as the view transformer to elevate $f_{i}$ into 3D voxel features $f_{v}\in \mathbb{R}^{D \times H_v \times W_v \times Z_v}$, incorporating discrete depth distributions $f_d$ and contextual features $f_{c}$ from geometric and contextual decoders, via Equation~\ref{eq:lss}:
\begin{equation}
f_v = \mathcal{F}_{lss}\left(f_{c}, f_{d}, \mathcal{K}^{lidar}_{cam}\right),
\label{eq:lss}
\end{equation}
where $\mathcal{K}^{lidar}_{cam}$ denotes calibrated parameters that facilitate the transformation from camera to LiDAR coordinates. 

Then, to perform 3D detection and occupancy (Occ.) prediction from bird’s-eye-view (BEV),  existing methods~\cite{huang2021bevdet, li2023bevdepth, li2022bevformer} typically collapse voxel features $f_v$ along the z-axis to form flattened global features $f_{g}$, inevitably overlooking fine-grained spatial structures of objects and resulting in suboptimal performance. To mitigate this issue, we adopt a voxel-height-guided sampling (VHS) strategy to capture local features $f_l$, preserving informative spatial information that reflects distinct object properties, as detailed in Sec.~\ref{subsec:VHS}. 

Furthermore, we empirically find that relying solely on task-specific local or global features, without effective and interactive feature fusion, hampers the scene understanding capacity of the 3D detection head. Therefore, we introduce a global-local collaborative feature fusion (CFF) mechanism in Sec.~\ref{subsec:GLCF} that integrates the complementary structural information from both global and local features, $f_{g}$ and $f_{l}$, to construct unified BEV representations $f_{u}^{bev}$. 
To ensure the effective propagation of informative spatial properties of objects, we re-elevate the unified BEV features $f_{u}^{bev}$ to voxel features $f_{u}^{vox}$ using a channel-to-height transformation~\cite{yu2023flashocc}. Following this, both $f_{u}^{bev}$ and $f_{u}^{vox}$ are forwarded to task-specific heads, performing 3D detection and Occ. prediction, respectively, as denoted by Equation~\ref{eq:mtllearner}:
\begin{equation}
\{\hat{B}, \hat{V} \} = \mathcal{F}_{cop}\left(f_{u}^{bev}, f_{u}^{vox}\right),
\label{eq:mtllearner}
\end{equation}
where $\hat{B}$ represents the detected bounding boxes, and $\hat{V} \in \mathbb{R}^{M \times H \times W \times Z}$ denotes semantic occupancy probabilities of voxels, with $M$ denoting the number of semantic labels, including the unoccupied voxels indicated as \textit{empty}.
Therefore, the overall learning objective is:
\begin{equation}
\min_{\Theta_{cop}} \sum_i^{N_i} \left[\mathcal{L}_{det}(B_i, \hat{B_i})+\beta\cdot\mathcal{L}_{occ}(V_i, \hat{V_i})\cdot\mathcal{W}_v^i\right],
\end{equation}
where $\beta$ serves as a temperature parameter to balance the learning progress between tasks, and $\mathcal{W}_v$ denotes the local-density matrix, as detailed in Sec.~\ref{subsec:OccupancyGeneration}.
\subsection{Local-Dentisy-aware Spatial Occupancy Generation} \label{subsec:OccupancyGeneration}
\begin{figure}[]
\begin{center}
\includegraphics[width=1.0\linewidth]{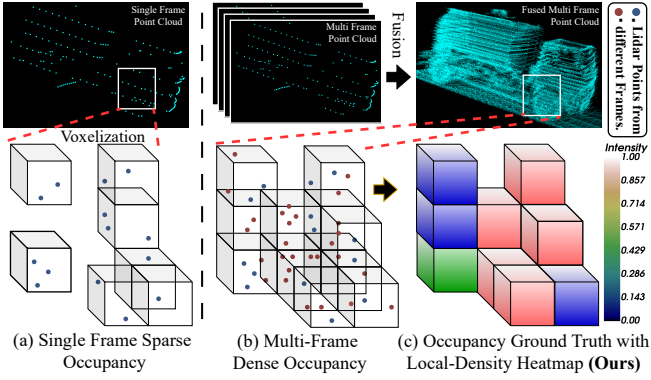}
\end{center}
\setlength{\abovecaptionskip}{1pt}
\caption{Illustration of different 3D Occupancy ground truth generation methods~\cite{huang2023tri,wei2023surroundocc}. Compared to others, our LDO pipeline introduces the local point density as prior and achieves a more detailed scene description.
}
\vspace{-4mm}
\label{fig:sparseDense}
\end{figure}

Given the inherent sparsity and varying densities of LiDAR data due to distance, natural divergence, and angular offsets between LiDAR sensors and targets, the voxelized 3D occupancy (Occ.) derived from these sparse point clouds lacks fine object details, as depicted in Figure~\ref{fig:sparseDense}(a). Meanwhile, many prior studies~\cite{tian2024occ3d, tang2024sparseocc, wei2023surroundocc, hou2024fastocc} in 3D Occ. assume a homogeneous local density, overlooking the non-uniform property of point clouds, which is a critical factor in representing fine-grained object structures.

To alleviate this, inspired by previous work~\cite{zuo2023pointocc, tian2024occ3d, he2022density}, we propose a novel \textbf{l}ocal-\textbf{d}ensity-aware spatial \textbf{o}ccupancy (LDO) pipeline to automatically generate semantically rich 3D Occ. with a voxel-level weighted matrix to indicate local densities. To ensure the quality of the generated dense occupancy ground truths (GTs), we incorporate the existing detection and segmentation labels. Initially, as illustrated in Figure~\ref{fig:overview}(a), we denote $\mathcal{P}^{N_s}$ as the point clouds spanning $N_s$ frames within a single scene. To construct their dense representation $\mathcal{P}_d$, we employ a semantic extractor to distinguish point clouds between static targets $\mathcal{P}_s^{N_s}$ and dynamic targets $\mathcal{P}_t^{N_s}$ in each frame of the same scene according to their annotations. Then, to simplify the aggregation, we transform both $\mathcal{P}_s^{N_s}$ and $\mathcal{P}_t^{N_s}$ into the world coordinate system, yielding $\mathcal{P}_{ws}^{N_s} $ for static and  $\mathcal{P}_{wt}^{N_s}$ for dynamic targets. Intuitively, for static targets $\mathcal{P}_{ws}^{N_s}$, we concatenate the $N_s$ sparse LiDAR sequences using world coordinates, forming dense point clouds $\mathcal{P}_{ws}^{d} = \left\{\mathcal{P}_{ws}^{N_s^1} \oplus \mathcal{P}_{ws}^{N_s^2}\oplus \dots \oplus \mathcal{P}_{ws}^{N_s^i}\right\}$. However, for dynamic targets $\mathcal{P}_t^{N_s}$, relying solely on coordinate information cannot ensure high-quality aggregation due to substantial positional shifts. To address this, we employ bounding box indices $i$ as auxiliary information to identify sparse point cloud data $\mathcal{P}_{{wt}_i}^{N_s}$ for each object. Then, we concatenate sparse point clouds for object by object and merge them to form a dense representation $\mathcal{P}_{wt}^d$ for all dynamic targets, as summarized by Equation~\ref{targetDENSE}:
\begin{equation} \label{targetDENSE}
\mathcal{P}_{wt}^{\text{d}} =  \bigcup_{k=1}^{M_s} \left(\mathcal{P}_{{wt}_k}^{N_s^1} \oplus \mathcal{P}_{{wt}_k}^{N_s^2}\oplus \dots \oplus \mathcal{P}_{{wt}_k}^{N_s^i}\right),
\end{equation}
where $\bigcup_{k=1}^{M_s}$ denotes the union operation and $M_s$ is the total number of dynamic objects in one scene. Next, both aggregated static targets $\mathcal{P}_{ws}^{\text{d}}$ and dynamic targets $\mathcal{P}_{wt}^{\text{d}}$ are transformed back into LiDAR coordinates based on their specific locations and ego-poses within the target frame, forming the dense point clouds $\mathcal{P}_{d}$.

Additionally, to derive dense 3D occupancy voxels $V_d \in \mathbb{R}^{H \times W \times Z}$, where $H$, $W$, and $Z$ denote the voxel divisions along each axis, one straightforward way is to assign voxels containing points as \textit{occupied} and assume a homogeneous local density among them. However, this uniformity overlooks differences in detailed features represented by local densities~\cite{he2022density}. To overcome this, we employ a voxel-based local-density assigner to calculate the local density matrix, representing this non-uniformity, as depicted on the right side of Figure~\ref{fig:overview}(a). Specifically, for the point cloud $\mathcal{P}_{d}^{k_i}$ within the $i$-th voxel in the $k$-th object, we calculate its local density factor as $\mathcal{W}_{d}^{k_i} = \frac{\mathcal{P}_{d}^{k_i}}{\sum_{j=1}^{n} \mathcal{P}_{d}^{k_j}}$, where $n$ is the total number of voxels associated with the $k$-th dynamic object. Hence, the overall local density matrix $\mathcal{W}_d \in \mathbb{R}^{H \times W \times Z}$ is expressed as $\mathcal{W}_d = \sum_{k=1}^{M_s} \left(\mathcal{S}_d^k \odot \left(\mathbb{I}_d^k \oplus \mathcal{W}_{d}^k\right)\right)$. Here, $\mathbb{I}_d$ and $\mathcal{S}_d$ represent the base matrix and a sparse matrix indicating \textit{empty} voxels with zeros; $\odot$ and $\oplus$ denote element-wise product and summation, respectively. Combining $\mathcal{W}_d$ with $\mathcal{V}_d$, our final local density-aware occupancy is $\mathcal{V}_{ld} \in \mathbb{R}^{H \times W \times Z \times 2}$. Thus, with the local density matrix as auxiliary supervision signals, LDO introduces fine-grained voxel-level local details with informative spatial structures of objects. All the generated LDO data can be found in VisionKG~\cite{yuan2024visionkg}.
\subsection{Voxel-Height-guided Sampling} \label{subsec:VHS}
\begin{figure}[ht]
\begin{center}
\includegraphics[width=1.0 \linewidth]{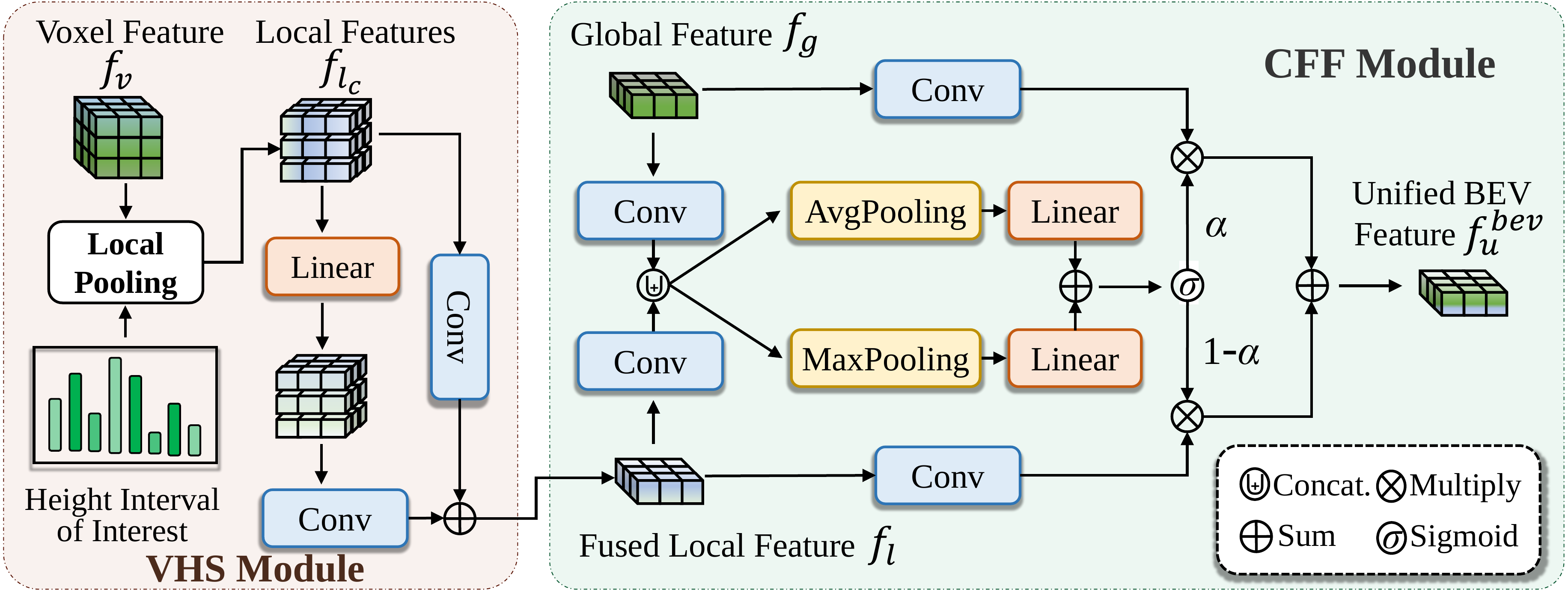}
\end{center}
\setlength{\abovecaptionskip}{1pt}
\caption{The proposed two-stage feature selection and fusion pipeline from CoP. The first stage (VHS) utilizes channel attention to merge and select multi-height local features. The second stage (CFF) serves to select features in the BEV space and explores both globally and locally distributed object properties.
}
\vspace{-5mm}
\label{fig:GLCF}
\end{figure}

BEV-based solutions~\cite{chi2023bev, li2023bevdepth, li2022bevformer} aggregate multi-view frustum features and collapse them along the z-axis to construct flattened global features $f_{g} \in \mathbb{R}^{C^{\prime}\times H^{\prime}\times W^{\prime}}$. However, this results in overlooking contextual semantics and preventing models from exploring the spatial properties of objects across diverse height ranges. To alleviate this, we employ a voxel-height-guided sampling (VHS) strategy to extract fine-grained local features $f_{l}  \in \mathbb{R}^{C^{\prime}\times H^{\prime}\times W^{\prime}}$, capturing their informative spatial structures. In contrast to the collapsed global features $f_g$, VHS involves hierarchical sampling guided by the height distribution of occupied voxels in the generated local density-aware occupancy $\mathcal{V}_{ld}$. Specifically, we employ voxel pooling~\cite{huang2021bevdet, li2023bevdepth, huang2022bevdet4d} over voxel features $f_{v}$ to extract the representative local features $f_{l_i}\in\mathbb{R}^{C^{\prime}\times H^{\prime}\times W^{\prime}}$, preserving the majority of local semantics, by Equation~\ref{eq:local}:
\begin{equation}
f_{l_i} = \textit{VoxelPooling}\left(f_{v}, H_{v}^{i}\right),
\label{eq:local}
\end{equation}
where $H_{v}^{i}$ denotes the specific height interval of interest (HoI) informed by occupied voxels.
These extracted local features $f_{l_i}$ are then concatenated to form $f_{l_c} \in \mathbb{R}^{L\times C^{\prime}\times H^{\prime}\times W^{\prime}}$, with $L$ representing  the number of voxel-informed HoIs. Subsequently, as illustrated in Figure~\ref{fig:GLCF}(a), we apply Squeeze-and-Excitation (SE) attention~\cite{hu2018squeeze} to adaptively aggregate these informative features. Specifically, the concatenated feature $f_{l_c}$ is transformed into $f_{l}^{{c_1}} \in \mathbb{R}^{C^{\prime} \times H^{\prime} \times W^{\prime}}$ via a $1\times 1$ convolution layer, reducing its channel dimension from $L \times C^{\prime}$ to $C^{\prime}$. Similarly, along a separate pathway, a linear layer followed by a $3\times 3$ convolution layer serves to reduce its channel dimensions to build $f_{l}^{{c_2}} \in \mathbb{R}^{C^{\prime} \times H^{\prime} \times W^{\prime}}$. Finally, the outputs from both pathways are summed to deliver fused local features $f_{l}$, setting the stage for subsequent feature fusion. Furthermore, unlike LiDAR-guided methods~\cite{duan2019centernet, chi2023bev} that utilize potentially redundant height information, our proposed VHS concentrates on the relative voxel heights, reducing noises from raw LiDAR data~\cite{ahmed2020density}. Benefiting from the LDO pipeline, the voxel-informed HoIs enable efficient pre-computation. 
\begin{table*}[ht]
\centering
\scalebox{0.95}{
\resizebox{1.0\textwidth}{!}{
\begin{tabular}{l|c|c|ccccc|c}
\toprule[1.0pt]

\textbf{Method} & \textbf{Modality} & \textbf{NDS} $\uparrow$ & \textbf{mATE} $\downarrow$ & \textbf{mASE} $\downarrow$ & \textbf{mAOE} $\downarrow$ & \textbf{mAVE} $\downarrow$ & \textbf{mAAE} $\downarrow$ & \textbf{mAP} $\uparrow$ \\
\hline
\hline
InfoFocus~\cite{wang2020infofocus}  & Lidar & 39.5 & 0.363 & 0.265 & 1.132 & 1.000 & 0.395 & 39.5 \\
PointPillars~\cite{lang2019pointpillars}  & Lidar & 45.3 & 0.517 & 0.290 & 0.500 & 0.316 & 0.368 & 30.5 \\
CenterFusion~\cite{nabati2021centerfusion}   & Lidar \& Radar &44.9 &0.326 &0.631 &0.261 &0.516 &0.614 &32.6 \\
MVFusion~\cite{wu2023mvfusion}   & Lidar \& Radar &51.7 &0.569 &0.246 &0.379 &0.781 &0.128 &45.3 \\
\hline
FCOS3D~\cite{wang2021fcos3d}  & Camera & 42.8 & 0.690 & 0.249 & 0.452 & 1.434 & 0.124 & 35.8 \\
DETR3D~\cite{wang2022detr3d}  & Camera & 47.9 & 0.641 & 0.255 & 0.394 & 0.845 & 0.133 & 41.2 \\
Ego3RT~\cite{lu2022learning}  & Camera  & 47.3 & 0.549 & 0.264 & 0.433 & 1.014 & 0.145 & 42.5 \\
BEVFormer~\cite{li2022bevformer}  & Camera & 53.5 & 0.631 & 0.257 & 0.405 & 0.435 & 0.143 & 44.5 \\
CMT-C~\cite{yan2023cross}  & Camera & 48.1 & 0.616 & 0.248 & 0.415 & 0.904 & 0.147 & 42.9 \\
BEVDet~\cite{huang2021bevdet}  & Camera & 48.2 & 0.529 & \textbf{0.236} & 0.395 & 0.979 & 0.152 & 42.2 \\
PETRv2~\cite{liu2023petrv2}  & Camera & 55.3 & 0.601 & 0.249 & 0.391 & 0.382 & \textbf{0.123} & 45.6 \\
X3KD~\cite{klingner2023x3kd}  & Camera & 56.1 & 0.506 & 0.253 & 0.414 & 0.366 & 0.131 & 45.6 \\
SOGDet~\cite{zhou2024sogdet} & Camera & \underline{58.1} & \textbf{0.471} & 0.246 & \underline{0.389} & \underline{0.330} & \underline{0.128} & \underline{47.4} \\
\hline
\hline
\rowcolor{gray!30} %
CoP (\textbf{Ours}) & Camera & \textbf{59.2} & \underline{0.482} & \underline{0.244} & \textbf{0.385} & \textbf{0.314} & 0.131 & \textbf{49.5} \\
\bottomrule[1.0pt]
\end{tabular}%
}
} 
\caption{Performance comparison on the nuScenes~\cite{caesar2020nuscenes} test set. Best number in \textbf{boldface} and \underline{underlined} number indicates the second-best performance.}
\label{tab:101_sota}
\end{table*}

\subsection{Global-Local Collaborative Feature Fusion} \label{subsec:GLCF}
Based on the extracted local and global features and inspired by~\cite{shi2024cobev, hou2024fastocc, yu2023flashocc}, we further employ a global-local collaborative feature fusion (CFF) strategy to achieve effective feature refinement, enhancement, and interaction. It leverages the synergistic cues from both extracted features and provides a seamless integration of complementary knowledge across tasks. To this end, we introduce parallel pathways to fuse the local $f_{l}$ and global $f_{g}$ features. As depicted in Figure~\ref{fig:GLCF}(b), our core concept involves selecting and combining the most relevant spatial and structural features from two heterogeneous, compressed feature sets. Initially, we incorporate average- and max-pooling operations, followed by two linear layers, to refine contextual information as described in  Equation~\ref{eq:ditill}:
\begin{equation}
f_{i}^{con}=\textit{MLP}\left(\mathcal{P}_{2d}\left(\textit{Conv}\left(f_{i}\right) \right)\right),
\label{eq:ditill}
\end{equation}
where $\mathcal{P}_{2d}$ denotes the pooling operation. 
Next, to ensure that complementary and representative features can freely flow in all dimensions during the fusion process, the adaptive selection parameter $\alpha$ is calculated as shown in Equation~\ref{eq:para}:
\begin{equation}
\alpha=\sigma\left(f^{con}_l \oplus f^{con}_g\right),
\label{eq:para}
\end{equation}
where $\oplus$ denotes element-wise summation and $\sigma\left(\cdot\right)$ represents the sigmoid activation function. $f^{con}_l$ and $f^{con}_g$ correspond to the refined local and global contextual features. Then, unified BEV features $f_{u}^{bev}$ can be obtained as follows: 
\begin{equation}
f_{u}^{bev}=\alpha \odot \textit{Conv}\left(f_{g}\right)+\left(1-\alpha\right) \odot \textit{Conv}\left(f_{l}\right),
\end{equation}
where $\odot$ denotes an element-wise product. This feature selection mechanism operates on the global-local spatial manifold, effectively squeezing each feature into a matrix with scalar values $\alpha 
\in \left(0,1\right)$. It enables the proposed CFF to perform a nuanced selection through weighted averaging, emphasizing the learning of both the globally distributed object properties and locally sensitive granularity. \\
Furthermore, structural information regarding vertical positioning is a key factor in ensuring the quality of spatial occupancy prediction. To perform effective propagation, drawing on FlashOcc~\cite{yu2023flashocc}, we utilize a channel-to-height (C2H) module to transform the unified BEV features $f_{u}^{bev} \in \mathbb{R}^{C^{\prime}\times H^{\prime}\times W^{\prime}}$ back into spatial voxel representations $f_{u}^{vox} \in  \mathbb{R}^{C^{\prime\prime} \times Z \times H \times W}$ with $C^{\prime} = C^{\prime\prime} \times Z$ as shown in  Equation~\ref{eq:c2h}:
\begin{equation}
\left\{f_{1}^{vox}, \ldots, f_n^{vox}\right\}=\text {C} 2 \textit {H}\left(\left\{f_{1}^{bev}, \ldots, f_n^{bev}\right\}\right) \text {, }
\label{eq:c2h}
\end{equation}
where $n$ denotes the number of predefined grids in BEV space.
In this way, by leveraging complementary cues and structural similarities between the detection and occupancy tasks, CoP is guided to distill the representative fused features in a collaborative learning manner.
\section{Experiments}
\subsection{Experimental Setups}
\noindent\textbf{Datasets and Metrics.} To assess the performance of the proposed CoP, we conduct extensive experiments on  nuScenes~\cite{caesar2020nuscenes}, a large-scale autonomous driving dataset as a primary benchmark for 3D object detection and occupancy prediction. It comprises 28,130 training samples, 6,019 validation samples, and 6,008 for evaluation. For 3D detection, we adopt the official evaluation metrics, including mAP, mATE, mASE, mAOE, mAVE, mAAE, and NDS, following previous works~\cite{huang2021bevdet,li2022bevformer, li2024dualbev, li2023bevdepth, huang2022bevdet4d}, providing a comprehensive evaluation involving center distance, translation, scale, orientation, velocity, and attribute. For 3D occupancy prediction, we report both scene completion IoU (SC. IoU) and semantic scene completion mIoU (SSC. mIoU) for a fair comparison. \\
\noindent\textbf{Implementation Details.} The proposed CoP is built upon BEVDet4D~\cite{huang2022bevdet4d} incorporating an effective forward projection via LSS~\cite{philion2020lift} and a camera-aware depth network from BEVDepth~\cite{li2023bevdepth}. Specifically, we discretize the depth axis into 60 points and define the spatial region as \([-51.2\text{m}, 51.2\text{m}]\) along the X and Y axes, and \([-3\text{m}, 5\text{m}]\) along the Z axis. We configure the AdamW~\cite{loshchilov2017decoupled} optimizer with a weight decay of 0.01 and set the initial learning rate to \(2 \times 10^{-4}\). To ensure a fair comparison with leading detectors~\cite{wang2022detr3d, huang2021bevdet, li2023bevdepth, li2022bevformer, liu2023petrv2, zhou2024sogdet}, we adopt ResNet-50 (R50) and ResNet-101 (R101)~\cite{he2016deep} as our image backbones and employ the CBGS~\cite{zhu2019class} strategy to mitigate class imbalance. Additionally, all experiments are trained for 24 epochs using four Nvidia A100 GPUs, each processing a batch size of four.
\begin{table}[ht]
\centering
\begin{tabular}{l|c|c|ccc}
\toprule[1.0pt]
Method & Venue & Input Size & NDS $\uparrow$ & mAP $\uparrow$ \\
\hline
BEVDet~\cite{huang2021bevdet} & arXiv22 &256 $\times$ 704 & 37.9 & 29.8 \\
Ego3RT~\cite{lu2022learning} & ECCV22 &256 $\times$ 704 & 40.9 & 35.5 \\
PETR~\cite{liu2022petr} & ECCV22 &256 $\times$ 704 & 43.1 & 36.1 \\
BEVDet4D~\cite{huang2022bevdet4d} & arXiv22 &256 $\times$ 704 & 45.7 & 37.2 \\
BEVDepth~\cite{li2023bevdepth} & AAAI23 &256 $\times$ 704 & 47.5 & 35.1 \\
AeDet~\cite{feng2023aedet} & CVPR23 &256 $\times$ 704 & 50.1 & 38.2 \\
Dual-BEV~\cite{li2024dualbev} & ECCV24 &256 $\times$ 704 & 50.4 & 38.0 \\
IA-BEV~\cite{jiao2024instance} & AAAI24 &256 $\times$ 704 & 51.6 & 40.0 \\
\hline
CoP (\textbf{Ours}) & - &256 $\times$ 704 & \textbf{53.5} & \textbf{42.2} \\
\bottomrule[1.0pt]
\end{tabular}
\caption{Quantitative results on the nuScenes-val~\cite{caesar2020nuscenes}.}
\vspace{-5mm}
\label{tab:r50_sota}
\end{table}
\begin{figure*}[t]
\begin{center}
\includegraphics[height=2.2in]{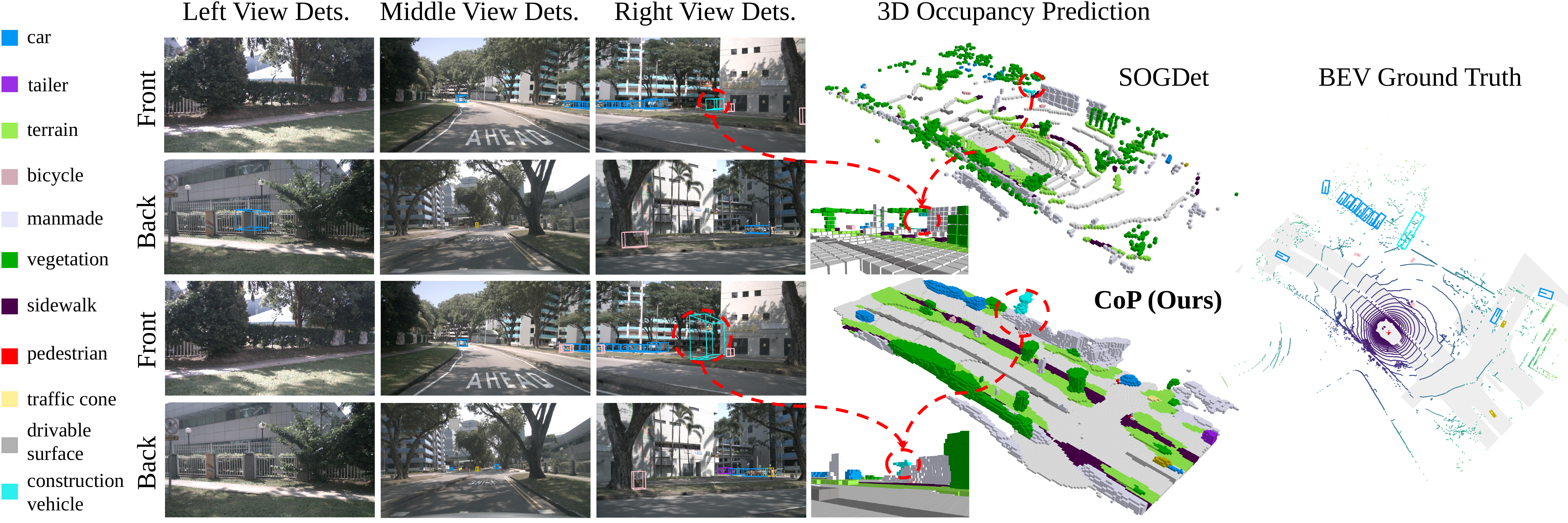}
\end{center}
\setlength{\abovecaptionskip}{1pt}
\vspace{-2mm}
\caption{
Qualitative results of the 3D detection and occupancy prediction on the validation set of nuScenes~\cite{caesar2020nuscenes}.
}
\vspace{-6mm}
\label{fig:comparison}
\end{figure*}
\subsection{Comparisons with State-of-the-art Methods} \label{subsec:sota}
As demonstrated in Table~\ref{tab:r50_sota}, we first evaluated CoP against leading 3D detectors, IA-BEV~\cite{jiao2024instance}, Dual-BEV~\cite{li2024dualbev}, and BEVDet4D~\cite{huang2022bevdet4d}, on the nuScenes validation set. It indicates that CoP gains the best performance with 53.5\% NDS and 42.2\% mAP, exceeding the robust baseline IA-BEV~\cite{jiao2024instance} by $+$1.9\% NDS and $+$2.2\% mAP. By enlarging the backbone to ResNet-101~\cite{he2016deep} and leveraging the enhanced capacity of CoP, we achieve impressive performance against advanced detectors~\cite{zhou2024sogdet,liu2023petrv2,wang2022detr3d,huang2021bevdet}, as shown in Table~\ref{tab:101_sota}. Specifically, our proposal achieves 59.2\% NDS and 49.5\% mAP on the nuScenes test set, marking an improvement of $+$1.1\% NSD and $+$2.1\% mAP over its rank-1 competitor~\cite{zhou2024sogdet}.

\begin{table}[ht]
\centering
\scalebox{0.85}{
\resizebox{0.56\textwidth}{!}{
\begin{tabular}{c|c|c|c|c|cc}
\toprule[1.0pt]
Sparse Occ. & Dense Occ. & LDO& VHS &CFF & NDS $\uparrow$ & mAP $\uparrow$ \\ \hline
\checkmark & & & & & 48.9 & 37.3 \\ \hline
& \checkmark & & & & 50.5 & 38.8 \\
& \checkmark & \checkmark & & & 50.7 & 39.2 \\
& \checkmark & \checkmark & \checkmark & & 52.3 & 40.6 \\ \hline
& \checkmark & \checkmark & \checkmark & \checkmark & 52.6 & 41.1 \\ 
\bottomrule[1.0pt]
\end{tabular}}}
\caption{Ablation study on nuScenes-val with R50~\cite{he2016deep}.}
\vspace{-2mm}
\label{tab:ablation}
\end{table} 
Furthermore, Figure~\ref{fig:comparison} demonstrates the performance of CoP in capturing diverse objects with both unique and irregular shapes. This stems from the global-local feature selection mechanism and the cross-task learning between 3D object detection and occupancy prediction, which are essential for allowing representative features to freely flow across both branches during the fusion process, thereby achieving geometric consistency and extracting valuable latent information. In contrast, SoGDet~\cite{zhou2024sogdet} either inaccurately falsely detects non-occupied positions as targets or overlooks the spatial structural consistency of construction vehicles. Meanwhile, Figure~\ref{fig:qualitativeResults} demonstrates that our method can also effectively capture geometric details across different height ranges and corresponding semantics in 3D occupancy prediction task. It also confirms the effectiveness of CoP in capturing both fine-grained and spatial structural information, as well as mining consistent knowledge from a joint perspective, suggesting that cross-task complementary and correlated cues are vital for models to perceive structural consistency and integrity in the physical world.
\subsection{Ablation Study} \label{subsec:ablation}
\noindent \textbf{Effects of Local-Density-Aware Dense Occupancy.}
As shown in Table~\ref{tab:ablation}, incorporating dense occupancy as supervision signals, CoP achieves a notable improvement in mAP by $+$1.5\% against the sparse model. This improvement is further amplified to $+$1.9\% with the proposed local-density-aware strategy. This suggests that local density supervision effectively alleviates spatial misalignment between the voxel features and the point cloud, reinforcing fine-grained feature consistency across both tasks. Meanwhile, we further examined its impact on various leading Occ. methods~\cite{huang2023tri, wei2023surroundocc}. Table~\ref{tab:ablationWeighting} confirms the scalability of the proposed LDO on strong baselines, TPVformer~\cite{huang2023tri} and SurroundOcc~\cite{wei2023surroundocc}, with mIoU increases by $+$0.3\% and $+$0.4\%, respectively, without an uptick in inference cost. Additionally, benefiting from LDO and collaborative learning, CoP achieved $+$0.7\% gain in IoU, with comparable speed to Surroundocc~\cite{wei2023surroundocc}. This finding establishes that the proposed LDO can serve as a versatile plug-and-play enhancement, bolstering existing Occ. solutions seamlessly, and substantially enhancing the details of voxel features.

\begin{table}[ht]
\centering
\scalebox{0.85}{
\resizebox{0.55\textwidth}{!}{
\begin{tabular}{l|cc|cc}
\toprule[1.0pt]  
Methods & SC. IoU & SSC. mIoU & Mem. (GB) & Latency (ms)  \\
\hline
TPVFormer~\cite{huang2023tri} & 30.9 & 17.1 & \multirow{2}{*}{4.68} & \multirow{2}{*}{371.4}\\
TPVFormer*~\cite{huang2023tri} & 31.2 & 17.3 &   & \\ \hline
SurroundOcc~\cite{wei2023surroundocc} & 31.5 & 20.3  & \multirow{2}{*}{5.71} & \multirow{2}{*}{513.6}\\
SurroundOcc*~\cite{wei2023surroundocc} & 31.9  & 20.7 &   &\\ \hline
CoP (\textbf{Ours}) & 32.6 & 20.9 & 4.69  & 509.6\\
\bottomrule[1.0pt]
\end{tabular}}}
\caption{Effect of the proposed local-density-aware occupancy and profiling on Nvidia-A40 with R101~\cite{he2016deep}(*indicates the models trained with our LDO strategy).}
\vspace{-7mm}
\label{tab:ablationWeighting}
\end{table}

\noindent\textbf{\textbf{\\ Effects of Voxel-Height-guided Sampling.}}
To assess the impact of the proposed voxel-height guided sampling strategy, we conduct experiments comparing it with both global and uniformly distributed height intervals, as shown in Tab~\ref{tab:ablationGridFlops}. It indicates that with uniformly distributed heights at intervals of \(2 \text{m}\), CoP yields a significant marginal increase in mAP by $+$1.1\%. This indicates height information not only promotes the 3D Occ. task, but also boosts detection task. Furthermore,  when voxel-height distributions serve as priors to distill local features, CoP yields a pronounced improvement $+$0.7\% increase in mAP,  by infusing structural semantics into the BEV representations. It indicates, with voxel-level height information as auxiliary signals, CoP can preserve informative
spatial information that reflects distinct object properties, as depicted in Figure~\ref{fig:comparison}.

\begin{table}[ht]
\centering
\scalebox{0.85}{
\resizebox{0.55\textwidth}{!}{
        \begin{tabular}{l|c|cc|c}
        \toprule[1.0pt]
        Sampling Strategy &mAP $\uparrow$ &mATE $\downarrow$ &mAVE $\downarrow$ & NDS $\uparrow$ \\
        \hline
        Global Pooling & 39.3 & 0.595& 0.349 & 50.6  \\
        Uniform Pooling & 40.4 & 0.585 & 0.349 & 50.9  \\ \hline
        w/ Local Pooling (BL) & 40.1 & 0.552 & 0.327 & 52.0  \\
        w/ Local Pooling (BL + UL)  & 40.8 & 0.553 & 0.316 & 52.6 \\
        w/ Local Pooling (BL + UL + EFL)  & 41.1 & 0.572 & 0.312 & 52.6 \\
        \bottomrule[1.0pt]
\end{tabular}}}
\caption{Ablation study of different sampling strategies.}
\vspace{-7mm}
\label{tab:ablationGridFlops}
\end{table}
\textbf{\textbf{\\ Effects of Global-Local Collaborative Feature Fusion.}}
As demonstrated in Table~\ref{tab:ablation}, with CFF, our proposed method yields 41.1\% mAP, an improvement of $+$3.8\% over the baseline, which relies exclusively on task-constrained features. This leverages cross-task spatial and structural similarities, to construct complementary and robust BEV representations from global and local features, surpassing strategies without adaptive feature selection. This suggests that detailed voxel semantic features can substantially enhance the performance of the detection head through scene-level fine-grained comprehension. Without adaptive feature selection, task heads may lack the capacity to extract consistent semantic information, leading to performance degradation. 
\vspace{-1mm}
\section{CONCLUSIONS}
\vspace{-1mm}
In this study, we introduce CoP, a robust cross-task framework that enhances multi-view 3D object detection by leveraging spatial consistency. To achieve this, we employ dense occupancy with detailed local-density-aware supervision to capture structural and conceptual similarities across tasks. Moreover, we implement a voxel-height-guided local sampling strategy to prevent the collapsing effects associated with flattened global BEV features, extracting fine-grained local features across diverse height ranges and object properties. Furthermore, our global-local collaborative feature fusion module facilitates the construction of robust BEV representations, ensuring seamless integration of complementary knowledge from both tasks. Overall, CoP attains state-of-the-art performance in 3D detection, with 59.2\% NDS and 49.5\% mAP on the nuScenes benchmark.

\section{Acknowledgements}
This work is supported by the Deutsche Forschungsgemeinschaft, German Research Foundation under grant number 453130567 (COSMO), by the Horizon Europe Research and Innovation Actions under grant number 101092908 (SmartEdge), by the Chips Joint Undertaking (JU), European Union (EU) HORIZON-JU-IA, under grant agreement No. 101140087 (SMARTY), by the Federal Ministry for Education and Research, Germany under grant number BIFOLD25B. 
We would also like to express our special thanks to Mr. Ali Ganbarov for his contributions in 3D occupancy processing during his academic stay at ODS, TU Berlin.



\bibliographystyle{IEEETran}
\bibliography{references}
\clearpage
\normalsize
\section*{Supplementary Materials}
In the following pages, we present experimental details, further analysis of the proposed method, and qualitative results.
\begin{bibunit}
    \section*{A. Details of Experimental Setup}
\subsection*{A.1. Datasets and Metrics}
\begin{figure}[h]
\begin{center}
\includegraphics[width=0.9\linewidth]{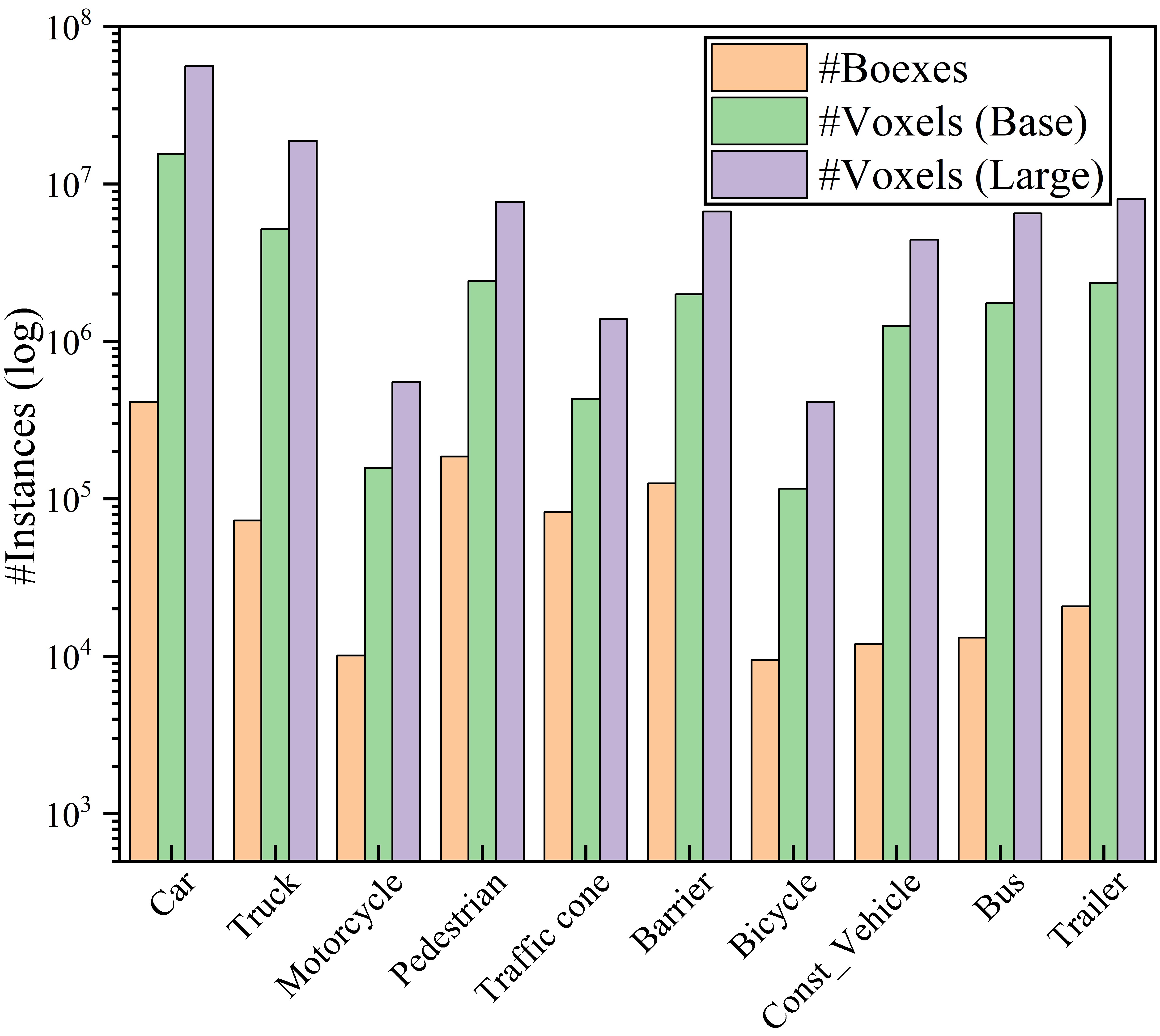}
\end{center}
\caption{
Class distribution of the nuScenes~\cite{caesar2020nuscenes} dataset used for 3D object detection and occupancy prediction.
}
\label{fig:figstat}
\end{figure}

We evaluate the proposed Collaborative Perceiver (CoP) on the NuScenes benchmark~\cite{caesar2020nuscenes}, which is currently the exclusive benchmark for both 3D object detection and occupancy prediction. The dataset was divided into training, validation, and evaluation sets, adhering to standard settings~\cite{caesar2020nuscenes}. For 3D detection, we adopted the official evaluation metrics, including mean Average Precision (mAP), mean Average Translation Error (mATE), mean Average Scale Error (mASE), mean Average Orientation Error (mAOE), mean Average Velocity Error (mAVE), mean Average Attribute Error (mAAE), and the representative metric NuScenes Detection Score (NDS). These metrics provide a comprehensive evaluation based on factors, e.g., center distance, translation, scale, orientation, velocity, and attribute. For 3D occupancy prediction, we report both scene completion IoU (SC. IoU) and semantic scene completion mIoU (SSC. mIoU) for a fair comparison: 1) SC. IoU measures voxel occupancy prediction accuracy for occluded voxels. It evaluates scene completion capability; 2) SSC. mIoU
measures the accuracy of predicting semantic labels for occupied voxels, where the per-class IoU, as well as the mean IoU (mIoU) across all classes, are reported. Notably, our local-density-aware dense occupancy ground truth is automatically generated without demanding human efforts during training. Overall dataset statistics used for 3D detection and occupancy tasks are summarized in Figure~\ref{fig:figstat}.

\subsection*{A.2. Implementation Details}
We implement the proposed CoP in PyTorch~\cite{paszke2019pytorch} using the MMDetection3D~\cite{mmdet3d2020} framework, with experiments primarily conducted on four Nvidia A100 GPUs. Adhering to standard evaluation settings~\cite{caesar2020nuscenes, li2022bevformer,zhou2024sogdet}, we conducted controlled experimental setups for equitable comparisons with prior work~\cite{huang2021bevdet,li2022bevformer, li2024dualbev, li2023bevdepth, huang2022bevdet4d}. We use pre-trained ResNet-50 (R50) and ResNet-101 (R101)~\cite{he2016deep} as the backbone to extract features from multi-view images. In the cross-task collaborative learning pathways of CoP, a Global-Local collaborative feature Fusion (GLF) module was applied, aiming to extract unified features for both 3D object detection and occupancy prediction. Different from prior solutions~\cite{huang2023tri, wei2023surroundocc}, we propose a Local-Density-aware dense Occupancy generation pipeline (LDO) to explicitly indicate the voxel-level points densities with a weighted matrix to introduce fine-grained voxel-level local details with
informative spatial structures of objects. 
To further boost models in exploring the spatial properties of objects across diverse height ranges and mitigate the overlooking contextual semantics caused by limited global heights, we employ a Voxel-Height-guided Sampling (VHS) strategy, which includes voxel pooling and local feature aggregation according to different height interval of interest (HoI). During both training and testing, we reduce the image resolution from the original $1600\times900$ pixels to $704\times256$ (R50) and $1408\times512$ (R101) pixels. Data augmentation is used during training, with random right-left flipping (with a probability of \(0.5\) and random resizing (from \(-6\) to \(11\) percent of image size). At testing time, we only use random resizing test augmentation where multi-view images with their resized version are fed into the network to decode the 3D bounding boxes and occupied probability with semantic predictions. Table~\ref{tab:hyperm} summarizes the full list of implementation details when using  R50 and R101~\cite{he2016deep} as the backbone network, respectively. In addition, we will release the code after publication.
\begin{table*}[ht]
\centering
\resizebox{0.8\textwidth}{!}{
\begin{tabular}{cccc}
\toprule[1.0pt]
Hyperparameter & Description & CoP (Base) & CoP (Large) \\
\midrule
- & Model backbone & ResNet-50~\cite{he2016deep} & ResNet-101~\cite{he2016deep} \\
- & Pretrained dataset & ImageNet~\cite{deng2009imagenet} & nuImg~\cite{caesar2020nuscenes} \\
- & Input image size & $256 \times 704$ & $512 \times 1408$ \\
$O_r$ & Resolution of the occupancy grid & $\left(0.8,0.8,0.8\right)$ & $\left(0.4,0.4,0.5\right)$ \\
$H_v$ & Height of Interest & \multicolumn{2}{l}{$\left[-3,-2\right], \left[-2,-1\right], \left[-1,0\right], \left[0,2\right], \left[-5,3\right], \left[-4,2\right], \left[-6,-4\right], \left[-2,1\right]$} \\
$D_d$ & Depth Discretization & \multicolumn{2}{c}{$\left[1.0, 60.0, 1.0\right]$} \\
$D_r$ & Distance Range & \multicolumn{2}{c}{$\left[-51.2, -51.2, -5.0, 51.2, 51.2, 3.0\right]$} \\
$\beta$ & Temperature Parameter & 0.9 & 0.9 \\
$b_s$ & Number of samples per batch & 16 & 16 \\
- & Optimizer & AdamW~\cite{loshchilov2017decoupled} & AdamW~\cite{loshchilov2017decoupled} \\
$W_d$ & Weight decay parameter & 1e-2 & 1e-2 \\
$l_r$ & Initial learning rate & 2e-4 & 2e-4 \\
- & Number of training epochs & 24 & 24 \\
- & Class-balanced group sampling~\cite{zhu2019class} & True & True \\
\bottomrule[1.0pt]
\end{tabular}
}
\caption{Model Hyperparameters and Experimental setup on the nuScenes~\cite{caesar2020nuscenes} Benchmark.}
\label{tab:hyperm}
\end{table*}

    \section*{B. Further Analysis}

\begin{table}[ht]
\centering
\scalebox{0.85}{
\resizebox{0.55\textwidth}{!}{
\begin{tabular}{l|cccc}
\toprule[1.0pt]
Voxel Size &mAP $\uparrow$ &mATE $\downarrow$ &mAVE $\downarrow$ & NDS $\uparrow$ \\
\hline
Voxel-64 (tiny) & 38.5 & 0.579& 0.368 & 50.4  \\ 
Voxel-128 (base)  & 41.1 & 0.572 & 0.312 & 52.6 \\
Voxel-256 (large) & 41.5 & 0.551 & 0.310 & 53.0 \\
\bottomrule[1.0pt]
\end{tabular}}}
\caption{Ablation study of different voxel sizes on nuScenes~\cite{caesar2020nuscenes} validation set with ResNet-50~\cite{he2016deep}.}
\label{tab:voxelSize}
\end{table}
\subsection*{B.1. Local-Dentisy-aware Dense Occupancy Generation}
In Section~\ref{subsec:OccupancyGeneration} of the manuscript, we elaborate on the LDO module, which aims to introduce the inherent sparsity and the non-uniform density of points across different distances and angles into generated occupancy GTs as priors, enhancing fine-grained voxel-level local details with informative spatial structures of objects. Beyond this, the granularity of the generated voxel sizes can also significantly affect the model's performance. Table~\ref{tab:voxelSize} shows that the coarsest voxel sizes \([64, 64, 5]\) result in lower mAP and NDS values compared to larger voxel sizes. This suggests that low voxel resolution may introduce noise and fail to provide consistent task representation for 3D detection and occupancy prediction, impairing the model's unified feature extraction. Conversely, the finest voxel resolution \([256, 256, 16]\) achieves the best performance, with 41.5\% mAP and 53.0\% NDS, indicating that fine-grained voxelization significantly improves the detection head's performance by enabling a detailed scene understanding. However,  compared to the base size \([128, 128, 10]\), it only improves mATE by $+$0.021 and mAVE by $+$0.002 with significantly higher computational load, as demonstrated in Figure~\ref{fig:figstat}. Therefore, to balance detailed scene analysis and computational efficiency, we employ the \textit{voxel-128 (base)} in our experiments for CoP (base), as demonstrated in Table~\ref{tab:hyperm}. We compare its performance to other models~\cite{li2024dualbev, jiao2024instance, feng2023aedet, li2023bevdepth} using ResNet-50, as shown in Table~\ref{tab:r50_sota}. It indicates that CoP gains the best performance with 52.6\%/41.1\% mAP/NDS, exceeding the robust baseline IA-BEV~\cite{jiao2024instance} by $+$1.1\%/$+$1.0\% NDS/mAP.
\begin{figure}[h]
\begin{center}
\includegraphics[width=0.9\linewidth]{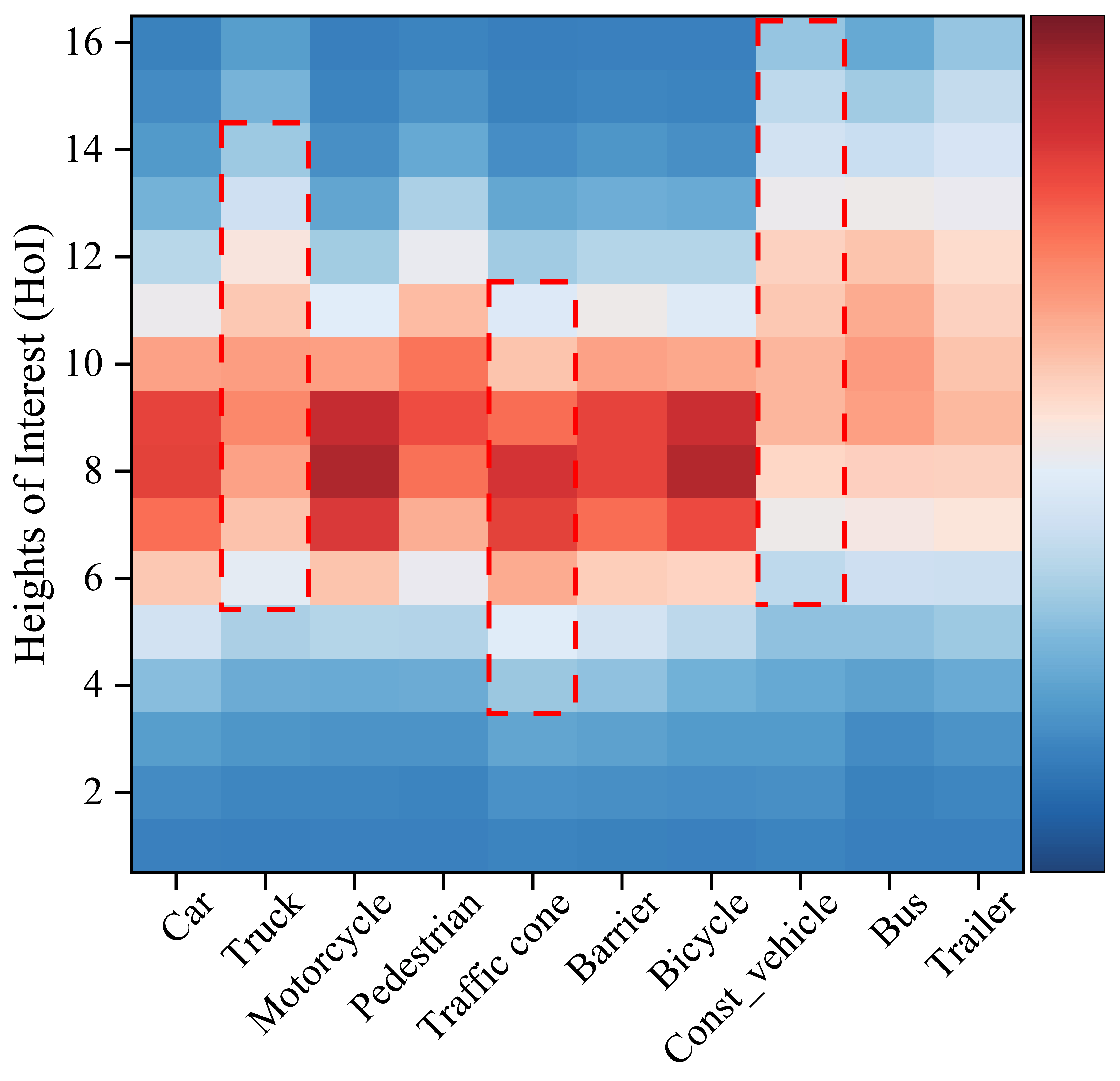}
\end{center}
\setlength{\abovecaptionskip}{1pt}
\caption{
Height distribution of the generated occupied voxels (large) from nuScenes~\cite{caesar2020nuscenes} dataset (each interval represents 0.5\text{m} in the real world).
}
\label{fig:figheat}
\end{figure}
\subsection*{B.2. Voxel-Height-guided Sampling}
To improve the effectiveness and robustness of the proposed CoP, VHS introduces the spatial properties of diverse objects across various height ranges within complex urban environments. According to Figure~\ref{fig:figheat}, during the voxel-height-guided sampling process, we divided the HoIs into three distinct layers: Base Layer (BL), Universal Layer (UL), and Extended Focus Layer (EFL), ensuring comprehensive coverage across various object types and sizes. Specifically, the BL layer spans heights in \([-3\text{m}, -2\text{m}]\), \([-2\text{m}, -1\text{m}]\), \([-1\text{m}, 0\text{m}]\) and \([0\text{m}, 2\text{m}]\), encompassing low-height located objects, such as roads, traffic cones, and vegetation. Next, the UL layer includes height intervals in \([-2\text{m}, 1\text{m}]\) and  \([-5\text{m}, 3\text{m}]\), and accommodates the majority of traffic participants, including cars, pedestrians, trailers, and trucks. The EFL layer concentrates between heights in \([-4\text{m}, 2\text{m}]\) and \([-6\text{m}, 4\text{m}]\), extends the field-of-view within a wider range, focusing less common but significant entities such as construction vehicles. In addition, we report the results of CoP with various local sampling strategies on nuScenes~\cite{caesar2020nuscenes} validation set with ResNet-50~\cite{he2016deep}. As can be seen from Table~\ref{tab:heightInterval}, our hierarchically local sampling strategy within the base layer can improve the baseline method by $+$1.8\%/$+$2.0\% in mAP/NDS.

\begin{table}[ht]
\centering
\scalebox{0.85}{
\resizebox{0.55\textwidth}{!}{
\begin{tabular}{l|cccc}
\toprule[1.0pt]
Sampling Strategy &mAP $\uparrow$ &mATE $\downarrow$ &mAVE $\downarrow$ & NDS $\uparrow$ \\
\hline
Global Pooling & 39.3 & 0.595& 0.349 & 50.6  \\ \hline
w/ Local Pooling (BL) & 40.1 & 0.552 & 0.327 & 52.0  \\
w/ Local Pooling (BL + UL)  & 40.8 & 0.553 & 0.316 & 52.6 \\
w/ Local Pooling (BL + UL + EFL)  & 41.1 & 0.572 & 0.312 & 52.6 \\
\bottomrule[1.0pt]
\end{tabular}}}
\caption{Ablation study of different sampling strategies on nuScenes~\cite{caesar2020nuscenes} validation set with ResNet-50~\cite{he2016deep}.}
\label{tab:heightInterval}
\end{table}
    \section*{C. Qualitative Results}
\begin{figure*}[t]
\begin{center}
\includegraphics[width=\linewidth]{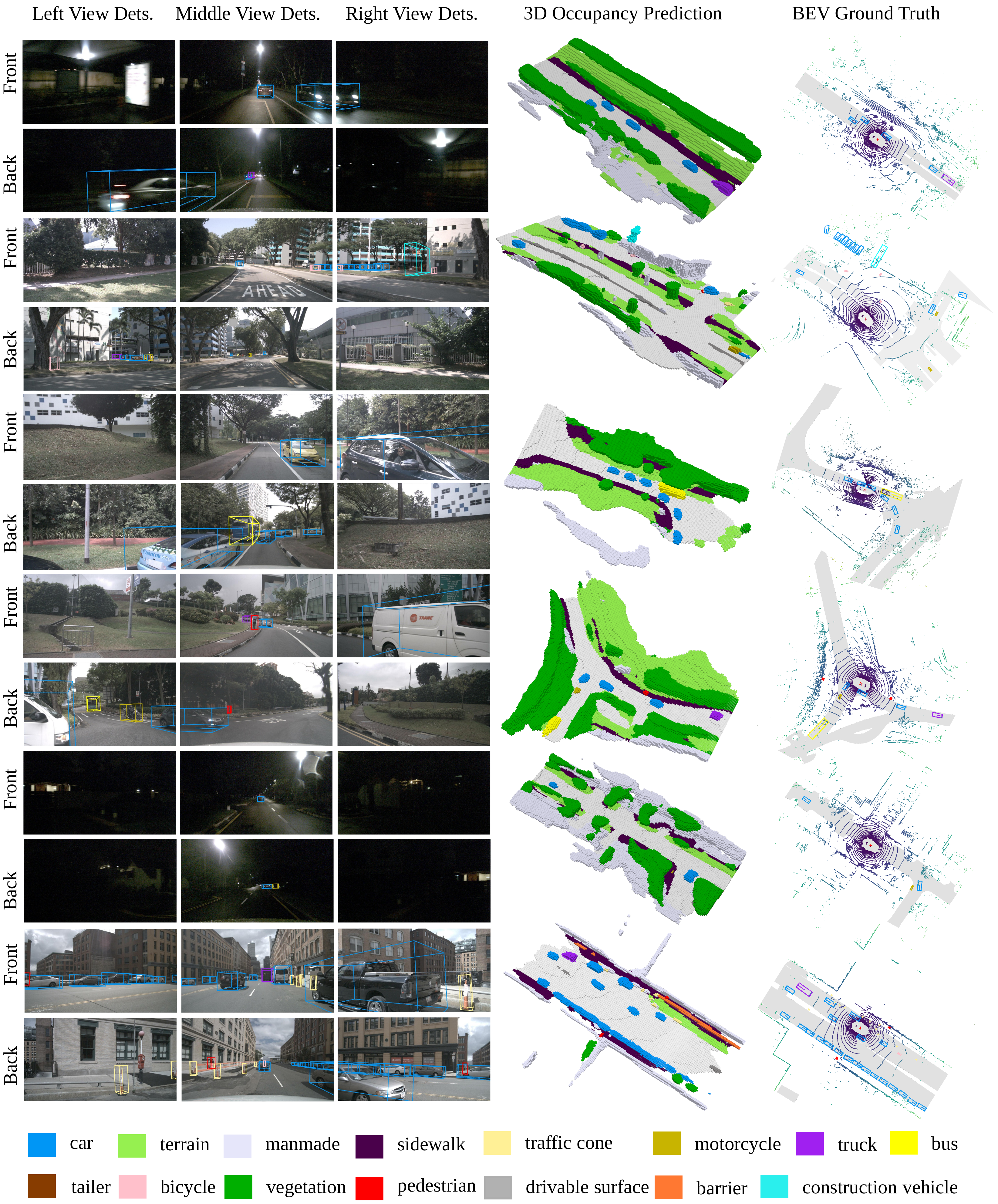}
\end{center}
\caption{Qualitative results of CoP (large) for the nuScenes~\cite{caesar2020nuscenes} dataset are displayed from left to right, showcasing 3D detection results, followed by 3D occupancy predictions, and BEV ground truth presented on the far right. To facilitate enhanced visualization of targets across multiple views, the images from the back left and back right cameras have been interchanged to maintain consistency in viewpoint orientation. 
}
\label{fig:qualitativeResults}
\end{figure*}

In this section, we present the qualitative results of CoP on the nuScenes~\cite{caesar2020nuscenes} towards tasks in both 3D object detection and occupancy prediction, as shown in Figure~\ref{fig:qualitativeResults}. With the help of our dual path learning strategy, CoP achieves consistent performance in capturing diverse objects with either
unique or irregular shapes. Meanwhile, the multi-learning between object detection and occupancy prediction heads can efficiently decrease the detection of false positives and ghost objects, and more missed objects have been detected. Another most obvious and common improvement is that the locations and orientations of the bounding boxes are further refined, benefiting from consistently unified features and environmental correspondences from joint perspectives of both tasks.
\begin{figure*}[t]
\begin{center}
\includegraphics[width=\linewidth]{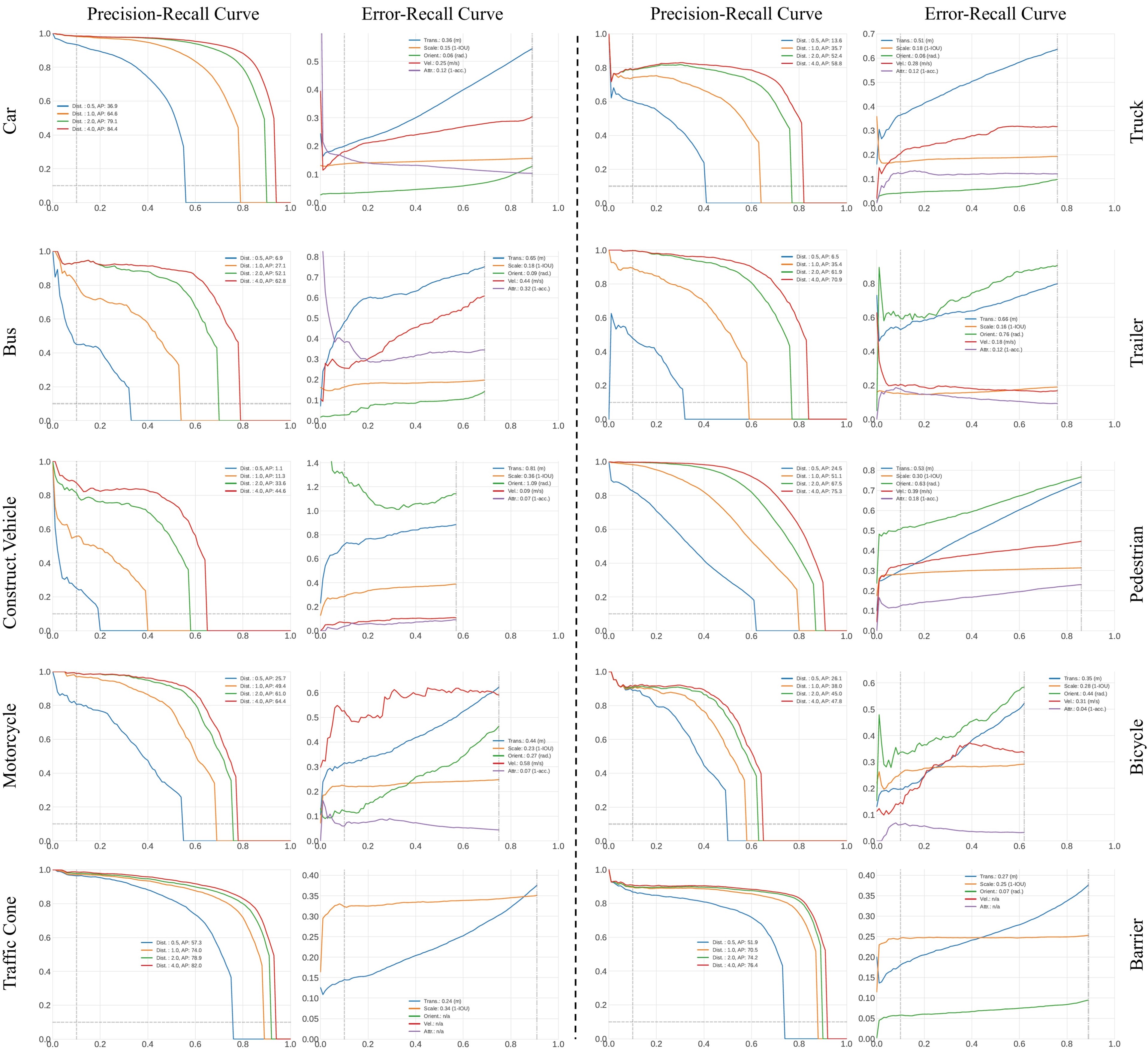}
\end{center}
\caption{
Class-wise Evaluation of CoP (large) on the test set of nuScenes~\cite{caesar2020nuscenes}.
}
\label{fig:nuBench}
\end{figure*}
\addtolength{\textheight}{-1.2cm}
    \putbib[supp]
\end{bibunit}
\end{document}